\documentclass[11pt]{article}

\usepackage[final]{acl}

\usepackage{times}
\usepackage{latexsym}

\usepackage[T1]{fontenc}

\usepackage[utf8]{inputenc}

\usepackage{microtype}

\usepackage{inconsolata}

\usepackage{graphicx}

\usepackage{enumitem}
\setlist[itemize]{leftmargin=\parindent,labelindent=\parindent,itemsep=1pt}

\usepackage{comment}

\usepackage{booktabs}
\usepackage{float}
\usepackage{amsmath}
\usepackage{hyperref}
\usepackage{fvextra}
\usepackage{framed}

\newcommand{\ptico}{\raisebox{-0.2ex}{\includegraphics[height=2ex]{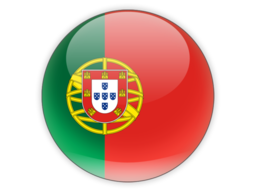}}}

\newcommand{\brico}{\raisebox{-0.2ex}{\includegraphics[height=2ex]{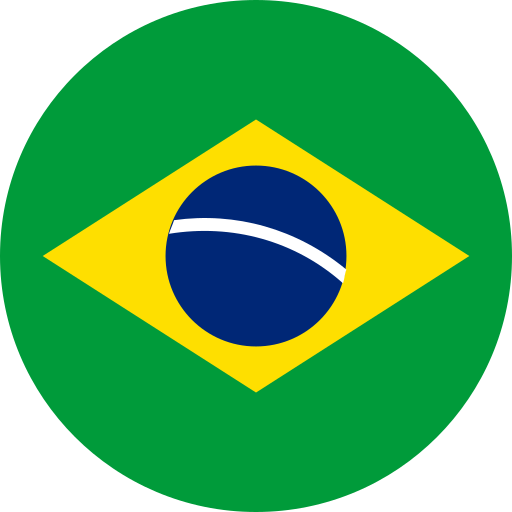}}}

\title{{\sc Math-PT}: \\ A Math Reasoning Benchmark for European and Brazilian Portuguese}

\author{
    \textnormal{Tiago Teixeira\textsuperscript{1},}
    \textnormal{Ana Carolina Erthal\textsuperscript{2},}
    \textnormal{Juan Belieni\textsuperscript{2},}
    \textnormal{Beatriz Canaverde\textsuperscript{1,3},}
    \\
    \textnormal{Diego Mesquita\textsuperscript{2},}
    \textnormal{Miguel Faria\textsuperscript{3},}
    \textnormal{Eliezer de Souza da Silva\textsuperscript{4,5},}
    \textnormal{André F. T. Martins\textsuperscript{1, 3}}
    \vspace{6pt}
    \\
    \textsuperscript{1}Instituto Superior Técnico, Universidade de Lisboa, \,\,
    \textsuperscript{2}Fundação Getulio Vargas,
    \\
    \textsuperscript{3}Instituto de Telecomunicações,
    \textsuperscript{4}Universidade de Coimbra, CISUC/LASI, DEI\\
    \textsuperscript{5}Basque Center for Applied Mathematics
    \\
    \small{\{tiago.renou, beatriz.canaverde, miguel.faria, andre.t.martins\}@tecnico.ulisboa.pt},\\
    \small{\{acarolerthal, juanbelieni\}@gmail.com}, \small{diego.mesquita@fgv.br},
    \small{eliezer.silva@uc.pt}
 }

\begin{document}
\maketitle
\begin{abstract}
The use of large language models (LLMs) for complex mathematical reasoning is an emergent area of research, with fast progress in methods, models, and benchmark datasets. 
However, most mathematical reasoning evaluations exhibit a significant linguistic bias, with the vast majority of benchmark datasets being exclusively in English or (at best) translated from English. 
We address this limitation by introducing {\sc Math-PT}, a novel dataset comprising 1,729 mathematical problems written in European and Brazilian Portuguese. {\sc Math-PT} is curated from a variety of high-quality native sources, including mathematical Olympiads, competitions, and exams from Portugal and Brazil. We present a comprehensive benchmark of current state-of-the-art LLMs on {\sc Math-PT}, revealing that frontier reasoning models achieve strong performance in multiple choice questions compared to open weight models, but that their performance decreases for questions with figures or open-ended questions. 
To facilitate future research, we release the benchmark dataset%
\footnote{\url{https://huggingface.co/datasets/tiagoteixeira03/MATH-PT}} %
and model outputs.%
\footnote{\url{https://github.com/deep-spin/math-benchmark}} %
\end{abstract}

\section{Introduction}

Evaluating mathematical reasoning has become a key challenge in large language model (LLM) research, motivating a growing ecosystem of benchmarks such as \textsc{MATH} \cite{hendrycks2021math}, MathVista \cite{lu2023mathvista}, MathBench \cite{liu-etal-2024-mathbench}, and MathArena \cite{balunovic2025matharena}. However, these resources share a major limitation: they are almost exclusively written in English, and multilingual variants typically resort to  translations of English benchmarks   \cite{shi2023language,son-etal-2025-linguistic, wang2025polymath}. This creates a linguistic bias that obscures whether LLM mathematical proficiency genuinely transfers across languages. 

Existing Portuguese NLP benchmarks, on the other hand, cover a  range of tasks---general evaluation suites \cite{paula-etal-2024-evaluation, scalercio-etal-2025-evaluating, baucells-etal-2025-iberobench}, toxicity detection \cite{da-silva-oliveira-etal-2024-toxic}, hate speech explanation \cite{salles-etal-2025-hatebrxplain}, truthfulness \cite{calvo-figueras-etal-2025-truth}, multi-factor explanations \cite{trager-etal-2025-mftcxplain}, and common-sense reasoning \cite{ponti-etal-2020-xcopa}. Yet, no benchmark targets mathematical reasoning, and none leverage the rich ecosystem of mathematical Olympiads and exams from Portugal and Brazil.

To address this gap, we introduce \textsc{Math-PT}, the first native Portuguese benchmark for mathematical reasoning. Our contributions are threefold:
\vspace{-0.1cm}
\begin{itemize}
    \item We introduce a new evaluation dataset natively in Portuguese, encompassing both the European and Brazilian variants. We curate \textbf{1,729 multiple-choice and open-ended math questions} from high-quality, originally Portuguese sources---including Olympiads and school exams from Portugal and Brazil, covering levels from primary to pre-university.
\vspace{-0.1cm}
    \item We benchmark \textbf{13 frontier and open-source LLMs}, with and without reasoning capabilities, revealing variations across dialects, difficulty levels, and problem types.
\vspace{-0.1cm}
    \item We show empirically that frontier models have strong performance on this task, but that accuracy varies significantly with \textbf{model size} and \textbf{question difficulty} (level, with/without figures, multiple choice/open ended), with significant room for improvement. 
\end{itemize}



\section{Dataset Creation}



\paragraph{European Portuguese \ptico} 
The questions in European Portuguese are sourced from the Portuguese Mathematical Olympiad (\textit{Olimpíadas Portuguesas da Matemática}, OPM), a yearly mathematics competition organized by \textit{Sociedade Portuguesa de Matemática} (SPM), which 
kindly provided the available \LaTeX{} files covering all editions of the competition from the 1997--98 to the 2024--25 academic years. 
The OPM are structured into four levels, each corresponding to specific school grades: \textbf{Level 1} (\textit{Pre-Olympiad}, 5\textsuperscript{th} grade), \textbf{Level 2} (\textit{Junior Category}, 6\textsuperscript{th}–7\textsuperscript{th} grades), \textbf{Level 3} (\textit{Category A}, 8\textsuperscript{th}–9\textsuperscript{th} grades), and \textbf{Level 4} (\textit{Category B}, 10\textsuperscript{th}–12\textsuperscript{th} grades). The Junior, A, and B categories each include two elimination rounds and a national final, whereas the \textit{Pre-Olympiad} consists of a single round designed to encourage participation in future editions. Each exam might feature \textbf{multiple-choice questions} (\textbf{MC}), \textbf{open-ended questions} (\textbf{OE}), or both. 
%
A non-negligible subset of the questions (316 out of 963) contains \textbf{visual elements}, particularly geometry-related problems. These figures are generated directly in \LaTeX{} using environments such as \texttt{picture}, \texttt{PStricks}, or \texttt{array}. For those questions, we preserve the original \LaTeX{} code and include it directly in the question to maintain the integrity of the visual information. Similarly, when a question statement contains tables (encoded using the \texttt{tabular} environment), we preserve the \LaTeX{} code, as this provides the most faithful and structured representation of tabular data for LLMs.

All mathematical expressions are kept in their original \LaTeX{} form, delimited by \texttt{\$\$}, following standard \LaTeX{} notation. All other \LaTeX{} markup was converted to plain text using the \texttt{pylatexenc} library, ensuring that the textual content remained readable while retaining mathematical fidelity. Statistics of the dataset are provided in Table~\ref{tab:questions-per-level}. 

\begin{table}[t]
\small
\centering
\setlength{\tabcolsep}{3pt} 
\begin{tabular}{llcrrr}
\toprule
\textbf{Country} & \textbf{Competition} & \textbf{Level} & \textbf{MC} & \textbf{OE} & \textbf{Total} \\
\midrule
\textbf{Portugal} & \textbf{OPM} & 1 & 78  & 34 & 112\\
                  &              & 2 & 169 & 96 & 265 \\
                  &              & 3 & 143 & 193 & 336 \\
                  &              & 4 & 29  & 221 & 250 \\
\midrule
\textbf{Brazil}   & \textbf{OBMEP} & 2 & 118 & -- & 118 \\
                  &                & 3 & 140 & -- & 140 \\
                  &                & 4 & 167 & -- & 167 \\
\cmidrule{2-6}
                  & \textbf{OBM}   & 2 & --  & 32 & 32\\
                  &                & 3 & --  & 58 & 58 \\
                  &                & 4 & --  & 38 & 38 \\
\cmidrule{2-6}
                  & \textbf{OMIF}  & 4 & 103 & -- & 103 \\
\cmidrule{2-6}
                  & \textbf{ELLM}  & 5 & 53  & -- & 53 \\
\cmidrule{2-6}
                  & \textbf{ITA}   & 5 & 57  & -- & 57 \\
\midrule
{\textbf{Total}} & & & \textbf{1,057} & \textbf{672} & \textbf{1,729} \\
\bottomrule
\end{tabular}
\caption{Number of questions per competition level and type across European and Brazilian Portuguese.}
\label{tab:questions-per-level}
\end{table}




\vspace{-0.2cm}
\paragraph{Brazilian Portuguese \brico} 
We collect questions from large-scale national competitions: the Brazilian Mathematical Olympiad of Public Schools (\textit{Olimpíada Brasileira de Matemática das Escolas Públicas}, OBMEP), the Mathematics Olympiad of Federal Institutions (\textit{Olimpíada de Matemática das Instituições Federais}, OMIF), the Elon Lages Lima Mathematics Competition (\textit{Competição Elon Lages Lima de Matemática}, ELLM) and the Entrance Exam for the Technological Institute of Aeronautics (\textit{Vestibular do Instituto Tecnológico de Aeronáutica}, ITA). 
The competitions are organized into distinct levels: \textbf{Level 2} (6\textsuperscript{th}–7\textsuperscript{th} grades), \textbf{Level 3} (8\textsuperscript{th}–9\textsuperscript{th} grades), \textbf{Level 4} (10\textsuperscript{th}–12\textsuperscript{th} grades) and \textbf{Level 5} (pre-university), where OBMEP covers levels 2--4, OMIF covers level 4, and both ELLM and ITA target level 5. Some competitions include both multiple-choice and open-ended exams. 

In all cases, \LaTeX{} sources are not publicly available, only PDF versions of the exams and official answer keys are distributed. Thus, we adopt an automatic extraction pipeline employing a smaller language model (\texttt{gpt-5-mini}) to read the PDF content and produce a structured representation of each exam. This approach was adopted due to the limitations of standard PDF-to-text conversion tools in accurately recovering the mathematical structures required for  question solving. 
Concretely, each exam PDF is first converted to page-level images using standard PDF tools. Then, each page is provided to \texttt{gpt-5-mini} with prompts instructing the model to (i) segment the page into individual questions; (ii) identify, for each question, the statement, the list of answer choices, and whether any visuals were necessary to solve the question; and (iii) normalize the output into a machine-readable schema (JSON). The prompts explicitly enforce constraints such as the expected number of options (typically five, labeled A--E), the need to represent all mathematical expressions in \LaTeX{} delimited by \texttt{\$\$} followed by standard \LaTeX{} notation, and the requirement that every extracted question be self-contained. 
Such competitions include questions that critically depend on visual information. In the PDFs, such content may appear as embedded raster images, vector diagrams, or tables. When the figure was provided through a clearly structured table (e.g., a small numeric grid or a simple schedule), \texttt{gpt-5-mini} was instructed to represent it using a minimal \texttt{tabular} environment, which we preserved in the final text. Questions with more complex diagrams and images, where visual content could not be reliably represented in plain text or \LaTeX{}, were discarded from the benchmark.








\section{Model Evaluation}



To ensure a consistent and reproducible evaluation protocol, we adopted a standardized prompting strategy adapted to each linguistic variant (European and Brazilian Portuguese) and to the presence or absence of visual elements in the question. 

For the multiple-choice questions, we use a direct instruction prompt in Portuguese (Figure~\ref{fig:ptpt-mlc-figure-prompt}). The prompt explicitly instructs the model to: (i) solve the multiple-choice mathematics question; (ii) output only the letter of the correct option (A, B, C, D, or E); and (iii) place that letter inside a \texttt{\textbackslash boxed\{\}} command. 
If the question contains figures (European Portuguese only), we  include an additional block enumerating the referenced visual content, which contains the \LaTeX{} source code with environments such as \texttt{picture}, \texttt{PStricks}, or \texttt{array} extracted from the source exam. 
For the Brazilian Portuguese subset (pt-BR), which consists exclusively of questions without figures, we adopt an analogous prompt written in Brazilian Portuguese. 
For open-ended questions, instead of asking for the letter of the correct option, we simply ask the model to make sure to place the final answer inside a \texttt{\textbackslash boxed \{\}} command (Figure~\ref{fig:ptpt-open-ended-prompt}).

\begin{figure}[t]
\scriptsize
\begin{framed}
\vspace{-0.1cm}
\begin{Verbatim}[breaklines=true, breaksymbol={}]
Resolve a seguinte questão de escolha múltipla de matemática. Certifica-te de colocar a letra da opção correta (A,B,C,D ou E), e apenas a letra da opção correta, dentro de \\boxed{}. Usa Português Europeu para pensar e responder.
Questão: Quantos fósforos são precisos para fomar o oitavo termo da seguinte sequência: (image1)
Conteúdo referenciado:
[image1]
\begin{picture}(...)\end{picture}
A) 21 B) 24 C) 27 D) 30 E) 34
\end{Verbatim}
\vspace{-0.5cm}
\end{framed}
\vspace{-0.2cm}
\caption{Example of a European Portuguese \ptico (pt-PT) prompt for a multiple-choice question with a figure.}
\label{fig:ptpt-mlc-figure-prompt}
\end{figure}
\begin{figure}[t]
\scriptsize
\begin{framed}
\vspace{-0.1cm}
\begin{Verbatim}[breaklines=true, breaksymbol={}]
Resolva a seguinte questão aberta de matemática. Certifique-se de colocar a resposta final dentro de \\boxed{}. Use Português do Brasil para pensar e responder.
Questão: Qual o menor valor de $n$ para que um polígono com $n$ lados tenha a soma de seus ângulos internos maior que $2012$ graus?
\end{Verbatim}
\vspace{-0.5cm}
\end{framed}
\vspace{-0.2cm}
\caption{Example of a Brazilian Portuguese \brico\, (pt-BR) prompt used for an open-ended question.}
\label{fig:ptpt-open-ended-prompt}
\end{figure}




All models are evaluated in a zero-shot setting, without chain-of-thought supervision or few-shot examples. Each question is submitted independently to the model using the appropriate prompt template. 
For multiple-choice questions, the model’s raw text output is parsed to extract the content inside the final \texttt{\textbackslash boxed\{\}} expression. A question is counted as correctly answered if and only if the extracted letter exactly matches the ground-truth answer key. 
For open-ended questions, we use Kimi K2 Thinking \cite{kimiteam2026kimik2openagentic} as an LLM judge to check if model answers are correct, i.e., mathematically equivalent to the golden answers. 
An example of a prompt sent to a judge is shown in Figure~\ref{fig:judge-prompt}.
Accuracy is computed separately for each competition level. 

\begin{figure}[t]
\scriptsize

\begin{framed}
\begin{Verbatim}[breaklines=true, breaksymbol={}]
 You are an impartial judge. Your task is to evaluate the correctness of a model's answer to an open-ended math question by comparing it ONLY to the gold solution.
 The model’s answer may use different reasoning steps or formatting than the gold solution; this is acceptable **as long as the final mathematical result is equivalent**. Focus strictly on mathematical correctness. Instructions:
 1. Compare the **final result** of the model answer to the gold answer.
 2. Minor algebraic, arithmetic, or formatting differences are allowed.
 3. Completely ignore writing style, explanation detail, or formatting.
 4. If the final results match or are mathematically equivalent → score 1.
 5. If the final results do not match → score 0.
 6. Output ONLY a JSON object in this exact format: {"score": 0 or 1, "explanation": "short explanation of the comparison"}
 [BEGIN DATA]
 ************
 [Golden Answer]: O preço dos $72$ pintainhos será representado por (...), pelo que cada pintainho foi vendido a $511$ escudos.
 ************
 [Model Answer]: Para resolver o problema, precisamos (...) Conclusão, o preço por pintainho no ano anterior era: $$\boxed{511}$$
 ************
 [END DATA]
 Provide your judgment now.
\end{Verbatim}
\vspace{-0.7cm}
\end{framed}
\vspace{-0.3cm}
\caption{Example of a prompt used for judging a model's answer to an open-ended question
}
\label{fig:judge-prompt}
\vspace{-0.4cm}
\end{figure}

The evaluated models include both closed models (GPT-5, Gemini 2.5 Flash, Claude Haiku 4.5) as well as open models (Qwen3, Gemma3, LLaMa3, DeepSeek Chat V3.1), 
covering a wide spectrum of model sizes and reasoning capabilities.

\begin{table}[t]
\centering
\scriptsize
\setlength{\tabcolsep}{3pt} 
\begin{tabular}{p{2.7cm} l c c c c}
\toprule
\textbf{Model} & \textbf{Level} & \multicolumn{2}{c}{pt-PT \ptico} & \multicolumn{2}{c}{pt-BR \brico} \\
\cmidrule(lr){3-4} \cmidrule(lr){5-6}
& & MC & OE & MC & OE \\
\midrule
\textbf{GPT-5} 
& L1 & \textbf{96.15} & \textbf{100.00} & N/A & N/A \\
& L2 & \textbf{94.67} & \textbf{89.58} & \textbf{90.68} & \textbf{90.62} \\
& L3 & \textbf{96.50} & \textbf{92.75} & \textbf{92.86} & \underline{91.38} \\
& L4 & \textbf{96.55} & \textbf{85.52} & \textbf{93.33} & \textbf{89.47} \\
& L5 & N/A & N/A & \textbf{90.00} & N/A \\

\midrule
\textbf{Qwen3-235B-A22B}
& L1 & \underline{91.03} & \underline{88.24} & N/A & N/A \\
& L2 & \underline{86.98} & \underline{79.17} & \underline{88.98} & \underline{84.38} \\
& L3 & 88.11 & \underline{81.35} & \textbf{92.86} & \textbf{93.10} \\
& L4 & \underline{93.10} & \underline{78.73} & \underline{91.48} & \underline{86.84} \\
& L5 & N/A & N/A & \underline{89.09} & N/A \\

\midrule
\textbf{Llama-3.3-70b-Instruct}
& L1 & 41.03 & 38.24 & N/A & N/A \\
& L2 & 32.54 & 17.71 & 35.59 & 34.38 \\
& L3 & 23.78 & 18.13 & 35.00 & 34.48 \\
& L4 & 24.14 & 20.36 & 32.96 & 18.42 \\
& L5 & N/A & N/A & 31.82 & N/A \\

\midrule
\textbf{Gemma-3-27b-it}
& L1 & 57.69 & 58.82 & N/A & N/A \\
& L2 & 49.11 & 30.21 & 66.95 & 65.62 \\
& L3 & 48.95 & 33.16 & 69.29 & 67.24 \\
& L4 & 58.62 & 27.60 & 64.81 & 52.63 \\
& L5 & N/A & N/A & 61.82 & N/A \\

\midrule
\textbf{Deepseek-chat-v3.1}
& L1 & 87.18 & 73.53 & N/A & N/A \\
& L2 & 84.02 & 67.71 & 83.05 & 78.12 \\
& L3 & \underline{88.81} & 72.02 & 91.43 & 86.21 \\
& L4 & 79.31 & 66.52 & 89.26 & 81.58 \\
& L5 & N/A & N/A & 86.36 & N/A \\

\midrule
\textbf{Claude-Haiku-4.5}
& L1 & 80.77 & 67.65 & N/A & N/A \\
& L2 & 74.56 & 47.92 & 83.90 & 68.75 \\
& L3 & 76.92 & 56.48 & 84.29 & 70.69 \\
& L4 & 68.97 & 47.51 & 81.48 & 60.53 \\
& L5 & N/A & N/A & 81.82 & N/A \\

\midrule
\textbf{Gemini-2.5-Flash}
& L1 & 83.33 & \underline{88.24} & N/A & N/A \\
& L2 & 75.74 & 72.92 & 83.90 & \underline{84.38} \\
& L3 & 83.92 & 74.61 & 88.57 & \underline{91.38} \\
& L4 & 79.31 & 71.04 & 85.93 & 71.05 \\
& L5 & N/A & N/A & \underline{89.09} & N/A \\
\bottomrule
\end{tabular}
\caption{Performance of several models in multiple choice and open-ended questions and different levels.}
\label{tab:acc-models}
\end{table}

\begin{table}[t]
\centering
\scriptsize
\setlength{\tabcolsep}{3pt}
\begin{tabular}{p{2.7cm} l c c c c}
\toprule
 \cmidrule(lr){3-6}
\textbf{Model} & \textbf{Level} & \multicolumn{2}{c}{MC} & \multicolumn{2}{c}{OE} \\
\cmidrule(lr){3-4} \cmidrule(lr){5-6}
& & No Fig. & Fig. & No Fig. & Fig. \\
\midrule
\textbf{GPT-5} 
& L1 & \textbf{100.00} & \textbf{88.00} & \textbf{100.00} & \textbf{100.00} \\
& L2 & \textbf{96.52} & \textbf{90.74} & \textbf{88.46} & \textbf{90.91} \\
& L3 & \textbf{97.92} & \textbf{93.62} & \textbf{94.35} & \textbf{89.86} \\
& L4 & \textbf{100.00} & \textbf{87.50} & \textbf{83.83} & \textbf{90.74} \\

\midrule
\textbf{Qwen3-235B-A22B} 
& L1 & \underline{98.11} & \underline{76.00} & \textbf{100.00} & 73.33 \\
& L2 & 90.43 & \underline{79.63} & \underline{82.69} & \underline{75.00} \\
& L3 & \underline{95.83} & 72.34 & \underline{86.29} & \underline{72.46} \\
& L4 & \underline{95.24} & \textbf{87.50} & \underline{80.24} & \underline{74.07} \\

\midrule
\textbf{Llama-3.3-70b-Instruct}
& L1 & 49.06 & 24.00 & 63.16 & 6.67 \\
& L2 & 40.00 & 16.67 & 21.15 & 13.64 \\
& L3 & 32.29 & 6.38 & 21.77 & 11.59 \\
& L4 & 28.57 & 12.50 & 19.76 & 22.22 \\

\midrule
\textbf{Gemma-3-27b-it}
& L1 & 66.04 & 40.00 & 73.68 & 40.00 \\
& L2 & 59.13 & 27.78 & 42.31 & 15.91 \\
& L3 & 58.33 & 29.79 & 33.87 & 31.88 \\
& L4 & 66.67 & 37.50 & 25.75 & 33.33 \\

\midrule
\textbf{Deepseek-chat-v3.1}
& L1 & \underline{98.11} & 64.00 & \textbf{100.00} & 40.00 \\
& L2 & \underline{94.78} & 61.11 & 73.08 & 61.36 \\
& L3 & 94.79 & 76.60 & 76.61 & 63.77 \\
& L4 & 80.95 & 75.00 & 68.26 & 61.11 \\

\midrule
\textbf{Claude-Haiku-4.5}
& L1 & 94.34 & 52.00 & 89.47 & 40.00 \\
& L2 & 84.35 & 53.70 & 57.69 & 36.36 \\
& L3 & 83.33 & 63.83 & 61.29 & 47.83 \\
& L4 & 76.19 & 50.00 & 47.90 & 46.30 \\

\midrule
\textbf{Gemini-2.5-Flash}
& L1 & 94.34 & 60.00 & 94.74 & \underline{80.00} \\
& L2 & 84.35 & 57.41 & 78.85 & 65.91 \\
& L3 & 86.46 & \underline{78.72} & 83.06 & 59.42 \\
& L4 & 80.95 & 75.00 & 71.86 & 68.52 \\
\bottomrule
\end{tabular}
\caption{Impact of visual content (figures) in pt-PT.}
\label{tab:acc-figs}
\end{table}

\begin{table}[t]
\centering
\scriptsize
\setlength{\tabcolsep}{3pt}
\begin{tabular}{p{2.7cm} l c c c c}
\toprule
\textbf{Model} & \textbf{Level} & \multicolumn{2}{c}{pt-PT \ptico} & \multicolumn{2}{c}{pt-BR \brico} \\
\cmidrule(lr){3-4} \cmidrule(lr){5-6}
& & MC & OE & MC & OE \\
\midrule
\textbf{Qwen3-8B} 
& L1 & 84.62 & \underline{82.35} & N/A & N/A \\
& L2 & 76.33 & 58.33 & 87.29 & 75.00 \\
& L3 & 78.32 & 61.14 & 87.14 & 72.41 \\
& L4 & 65.52 & 50.23 & 81.48 & 52.63 \\
& L5 & N/A & N/A & 76.36 & N/A \\

\midrule
\textbf{Qwen3-14B}
& L1 & 88.46 & \underline{82.35} & N/A & N/A \\
& L2 & \textbf{89.35} & 73.96 & \textbf{90.68} & 81.25 \\
& L3 & \underline{86.71} & \underline{76.68} & \underline{90.00} & \underline{84.48} \\
& L4 & 82.76 & 71.04 & 88.15 & 78.95 \\
& L5 & N/A & N/A & \textbf{90.91} & N/A \\

\midrule
\textbf{Qwen3-32B}
& L1 & 88.46 & 79.41 & N/A & N/A \\
& L2 & 83.43 & 70.83 & \underline{88.98} & \underline{84.38} \\
& L3 & 85.31 & 73.06 & 89.29 & 82.76 \\
& L4 & \textbf{93.10} & 65.16 & 85.19 & 76.32 \\
& L5 & N/A & N/A & 87.27 & N/A \\

\midrule
\textbf{Qwen3-30B-A3B}
& L1 & 82.05 & \underline{82.35} & N/A & N/A \\
& L2 & 82.84 & 64.58 & 82.20 & 75.00 \\
& L3 & \underline{86.71} & 63.73 & 89.29 & 74.14 \\
& L4 & 82.76 & 55.20 & 84.44 & 50.00 \\
& L5 & N/A & N/A & 80.00 & N/A \\

\midrule
\textbf{Qwen3-235B-A22B}
& L1 & \textbf{91.03} & \textbf{88.24} & N/A & N/A \\
& L2 & \underline{86.98} & \textbf{79.17} & \underline{88.98} & \textbf{84.38} \\
& L3 & \textbf{88.11} & \textbf{81.35} & \textbf{92.86} & \textbf{93.10} \\
& L4 & \textbf{93.10} & \textbf{78.73} & \textbf{91.48} & \textbf{86.84} \\
& L5 & N/A & N/A & \underline{89.09} & N/A \\
\bottomrule
\end{tabular}
\caption{Impact of model size within the Qwen3 family.}
\label{tab:acc-params}
\end{table}

\section{Analysis}




The results in \autoref{tab:acc-models} show consistent performance patterns across models, levels, and question types. Frontier models such as GPT-5, Qwen3-235B, Gemini 2.5 Flash, and Claude Haiku 4.5 outperform all mid-sized and open-weight models. GPT-5 is the strongest model overall, achieving the highest accuracy across nearly every setting in both pt-PT and pt-BR. Open-source models exhibit wider variability, with the Qwen family showing notably strong scaling, while Gemma-3 and Llama-3.3 underperform significantly in mathematical reasoning. 
A consistent gap emerges between multiple-choice (MC) and open-ended (OE) performance. In general, all models achieve higher accuracy on MC questions, and the difference widens substantially at higher difficulty levels, often exceeding 10\textendash20 points. GPT\textendash 5 shows the smallest MC\textendash OE gap, reflecting more stable reasoning capabilities, whereas mid-range models display large drops, especially in pt-PT. \looseness=-1

\autoref{tab:acc-figs} shows that problems involving figures introduce a prominent difficulty spike. Across all models, performance deteriorates sharply on pt-PT items referencing figure code, with drops of 10--20 points even for frontier models and up to 56 points for weaker baselines. This degradation is more pronounced in OE questions, indicating that multimodal reasoning remains a key bottleneck despite the textual representation of figures. 
For weaker models, accuracy decreases systematically from Level~1 to Level~4 and~5. While GPT-5 and Qwen3-235B remain robust even at the highest difficulty levels, other models exhibit sharp declines. 

Some scaling trends are also noticeable for Qwen3, as shown in \autoref{tab:acc-params}. Performance increases smoothly from 8B to 14B, 32B, 30B-A3B, and 235B in nearly all dimensions of the table. Scaling benefits are especially pronounced in OE tasks, indicating that larger models gain disproportionately in reasoning-heavy settings. Notably, Qwen3-14B performs unexpectedly well on pt-BR MC tasks, sometimes reaching accuracy levels comparable to older proprietary frontier models.

Comparing model architectures, Qwen3-235B is the strongest open-weight model exhibiting performances on par with the closed models. DeepSeek-Chat-v3.1 performs competitively on MC but struggles with OE and figure-based items. Llama-3.3-70B and Gemma-3-27b-it show limited mathematical reasoning capabilities.


\section{Conclusion}

We introduced {\sc Math-PT}, the first natively curated benchmark for mathematical reasoning in European and Brazilian Portuguese, sourced from real Olympiads and national exams. We observe that frontier models perform strongly on multiple-choice questions, with substantial drop in performance for open-ended questions and those those involving figures. 
By releasing {\sc Math-PT} we aim to support reproducible evaluation and to encourage progress toward more capable Portuguese-based mathematical reasoning models.

\section*{Acknowledgments} 
We would like to thank José Mourão and António Salgueiro for their help in obtaining the {\LaTeX} source for the OPM exams and solutions. 
The IT team is supported by the Portuguese Recovery and Resilience Plan through project C645008882-00000055 (Center for Responsible AI), by the project DECOLLAGE (ERC-2022-CoG 101088763), by the AMALIA project under Measure RE-C05-i08, and by FCT/MECI through national funds and when applicable co-funded EU funds under 10.54499/UID/50008/2025: Instituto de Telecomunicações. ES acknowledge the support of ``la Caixa'' Foundation’s LCF/BQ/PI22/11910028 award, as well as funds by MICIU/AEI/10.13039/501100011033 and the BERC 2022-2025 program funded by the Basque Government. ES research work is funded by Portuguese national funds through FCT – Foundation for Science and Technology, I.P., within the scope of the research unit UID/00326 - Centre for Informatics and Systems of the University of Coimbra. DM, ACE, and JB acknowledge the support by the Fundação Carlos Chagas Filho de Amparo à Pesquisa do Estado do Rio de Janeiro (FAPERJ) (SEI-260003/020694/2025) and the Conselho Nacional de Desenvolvimento Científico e Tecnológico (CNPq) (305692/2025-9).

\bibliography{custom}




\end{document}